\documentclass{article}

\usepackage{arxiv}

\usepackage[utf8]{inputenc} 
\usepackage[T1]{fontenc}    
\usepackage{hyperref}       
\usepackage{url}            
\usepackage{booktabs}       
\usepackage{amsfonts}       
\usepackage{nicefrac}       
\usepackage{microtype}      
\usepackage{lipsum}
\usepackage{float}
\usepackage{graphicx}
\graphicspath{ {./images/} }

\title{Predicting the Popularity of Reddit Posts with AI}

\author{
 Juno Kim \\
  Henry M. Gunn High School\\
  Palo Alto, CA 94306 \\
  \texttt{jk28093@pausd.us} \\
}

\begin{document}
\maketitle
\begin{abstract}
\quad Social media creates crucial mass changes, as popular posts and opinions cast a significant influence on users’ decisions and thought processes. For example, the recent Reddit uprising inspired by r/wallstreetbets which had remarkable economic impact was started with a series of posts on the thread. The prediction of posts that may have a notable impact will allow for the preparation of possible following trends. This study aims to develop a machine learning model capable of accurately predicting the popularity of a Reddit post. Specifically, the model is predicting the number of upvotes a post will receive based on its textual content. 

\quad I experimented with three different models: a baseline linear regression model, a random forest regression model, and a neural network. I collected Reddit post data from an online data set and analyzed the model’s performance when trained on a single subreddit and a collection of subreddits. The results showed that the neural network model performed the best when the loss of the models were compared. With the use of a machine learning model to predict social trends through the reaction users have to post, a better picture of the near future can be envisioned.

\end{abstract}



\section{Introduction}
\quad Reddit is a social media platform where individuals join subreddit, which are communities loosely based around a topic, such as dance. In the subreddit, users can make posts with images and text which other users can react to with upvotes and downvotes, which are approval and disapproval respectively. With the growing influence of social media on societal decisions and mindsets, predicting which posts will gain a large reaction is an important task.

\quad This study is an attempt to predict the number of Reddit upvotes that a post will receive based on its textual content. I used public data of the top 1000 posts from 2500 subreddits. I then use three different machine learning models: a linear regression model, a random forest regression model, and a neural network model to evaluate the combined text of the title and the post caption. I compared the loss of each of these models when predicting the number of upvotes to determine which is most effective.

\section{Related Work}
\label{sec:headings}

\quad The prediction of Reddit upvotes on posts allows technical applications in the prediction of huge trends on social media, following the recent surge of r/wallstreetbets against Melvin Capital hedge fund over GME(gamestop). An example analyzed in close detail by Wolfsled et al. \cite{wolfsfeld 2013} is the ‘Arab Spring’, a huge progression of uprisings against oppressive government systems with an emphasis on social media, the platform used to organize protests and spread information freely. Prediction through the analysis of social media has been trending with the vast amount of data now available, and Schoen et al. \cite{schoen2013} explore this with the prediction of future events and development in various fields, such as politics, finance, and more. There has been further research done into which fields can be predicted with social media and the techniques used for them by Yu and Kak \cite{yu2012}.

\quad Recently, the importance of social media trends specifically on Reddit was shown with the investor movement started by the subreddit r/WallStreetBets. The investors started a short squeeze against major hedge funds, pushing stocks such as GME(Gamestop) from 5 USD in July 2020 to a maximum of 483 USD on 28 January 2021. This resulted in huge losses for the hedge funds and more impacts, as studied by Lyócsa et al \cite{lyocsa2021}. The use of social power and movements with economic impact for the Reddit event has also been analyzed by Di Muzio [6]. This research shows how social trends and movements have huge impacts from an economic standpoint, and can be applied to other fields as well. The people who participated in this wave were also analyzed by Hasso et al \cite{hasso2021}. obtaining results showing a wave of people drawn to investing through a social as opposed to capital motivated movement.

\quad The popularity of Reddit comments has been predicted by Deaton et al \cite{deaton2017}. with the use of several machine learning models, such as Linear SVM, Decision Tree, and Naive Bayes. The models were trained on 53M comments to predict the hotness of a comment using relative confidence scores. There have been multiple papers on predicting the popularity of Reddit posts, with multiple models by Rohlin \cite{rohlin2016}. This paper uses Bag of Words, TF-IDF(Term Frequency-Inverse Document Frequency), and LDA(Latent Dirichlet Allocation) trained on features extracted with Naive Bayes and SVM. Segall and Zamoschin \cite{segall2012} also experimented by using Multiclass Naive Bayes. Multiclass SVM, and Linear Regression. I will build off these previous models analyzed to create a model that can more accurately predict the popularity of a Reddit post.

\section{Dataset}
\label{sec:others}
\quad The dataset that was used for this study is the Reddit Top 2.5 Million[] dataset on Github, which is a compilation of the top 1000 all-time posts from the top 2500 subreddits as of August 2013. The data is divided into 1000 row csv files by subreddit with features such as the time created, the score(upvotes - downvotes received), the title, selftext, and more. I removed unnecessary features, and kept the following: 

\begin{itemize}
\item \emph{ups} - The number of upvotes received
\item \emph{title} - The title of the post
\item \emph{selftext} - The text content of the post
\end{itemize}

I then created an additional feature that would be analyzed:
\begin{itemize}
    \item \emph{combined} - The concatenated \emph{title} and \emph{selftext}
\end{itemize}
Subreddits are based around different topics, such as the r/acting subreddit which is located in the respective acting.csv file. 

\section{Methods}
\subsection{Metrics}
The metric used to evaluate every model is RMSE, defined as:

\begin{equation}
    RMSE=\sqrt{\frac{1}{n}\sum_{i=1}^{n}\left(\frac{d_{i}-f_{i}}{\sigma}\right)^{2}}
\end{equation}

Given the residuals - the error of each prediction - the standard deviation is taken, resulting in RMSE which essentially returns how much the model’s prediction is off by and is sought to be minimized.

\subsection{Data Preprocessing}
In order to evaluate the models, I preprocessed the data by cleaning the text. I removed stopwords, or unimportant words using regex and the NLTK stopword collection. I first used only one csv: acting.csv which contains the top 1000 posts of the r/acting subreddit. I repeated these methods upon a combined Pandas Dataframe that contained all 2500 csv files.

\subsection{Models}
\subsubsection{Baseline Linear Regression}
A Linear Regression linear model was used as the baseline for the experiment. The Linear Regression model had an ordinary least squares loss objective and calculated the intercept of the model. As a baseline model, the linear regression model indicates the existence of more robust algorithms. One algorithm that fits this description is the Random Forest Regression Algorithm.
\subsubsection{Random Forest Regression}
Random Forest Regression improves upon the baseline linear regression model with the use of decision trees as opposed to a line of best fit. Random Forest Regression models use multiple independent decision trees and average the results in order to produce the ending prediction. This allows the model to outperform an individual decision tree due to having multiple independent trees from which mistakes are protected against simply due to their numbers. In direct contrast to the linear regression model, the Random Forest Regression model is able to find nonlinear interactions between the text and the number of upvotes. However, deep learning also provides another way to perform this task. 
\subsubsection{Neural Network}
To improve the Random Forest Regression model, I experimented with a neural network model. Neural networks use several algorithms as ‘layers’ while mimicking the human brain in order to analyze an input and output a single prediction. The architecture of the model is shown below:

\begin{figure}[H]
    \caption{Layer structure of Neural Network model}
    \centering
        \includegraphics[width=8cm]{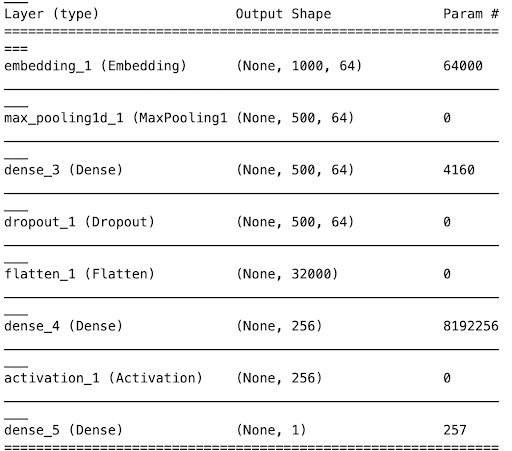}
\end{figure}

\section{Results}

\begin{table}[H]
    \caption{Table of RMSE Results for each Model on acting.csv and combined data}
  \centering
  \begin{tabular}{llll}
    \toprule               \\
    RMSE     & Linear Regression     & Random Forest & Neural Network \\
    \midrule
    Acting.csv & 178.66  & 13.22 & 8.10     \\
    Combined Data     & 402.61 & 405.19 & 394.26      \\
    \bottomrule
  \end{tabular}
  \label{tab:table}
\end{table}

\begin{figure}[H]
    \caption{Graph of RMSE vs. Model used on acting.csv}
    \centering
        \includegraphics[width=10cm]{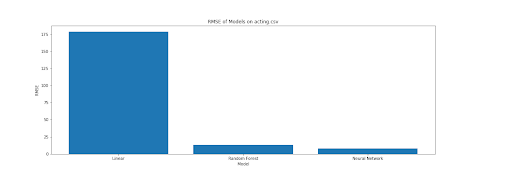}
\end{figure}

This graph shows the performance of the models on the acting.csv data, and it can be seen that the neural network model performed the best when evaluated on RMSE.

\begin{figure}[H]
    \caption{Graph of RMSE vs. Model used on combined data}
    \centering
        \includegraphics[width=10cm]{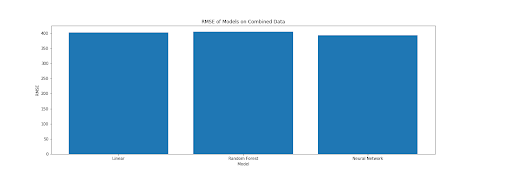}
\end{figure}

This graph shows the RMSE evaluation of the models on the combined data, and shows that the neural network performed the best, and the linear regression model performed better than the random forest regression model. 

In both instances, the Neural Network had the lowest loss and thus performed better than the other models. 

\section{Discussion}
\quad In order to accurately predict the popularity of a post I needed to decide which metric would best indicate this, between the number of upvotes received, the number of downvotes received, or the score, the number of downvotes subtracted from the number of upvotes. A high number of downvotes signals a post that a majority of users disagree or dislike. A high score indicates that more people agreed with the post than disagreed. However, popular posts with evenly split opinions will have a low score, in the same way that unpopular posts will have a low score due to no attraction. As such, I decided that using the number of upvotes was the best feature to predict as it indicates a popular post with the majority of users agreeing. 

\quad One outlier in the results is the RMSE of the Random Forest Regression model on the combined data as opposed to the linear regression model. This may have been caused by the fact that linear regression models can extrapolate while tree-based models like the Random Forest regression model cannot. However, this was most likely caused by irregular data, as the Random Forest regression model outperforms the linear regression model for the acting.csv showing that the selection of subreddits as data influences the performance of the models.

\section{Conclusion \& Future Work}
\quad In this paper, I used Reddit post data in order to evaluate three different models that predicted the upvotes of a Reddit post based on the combined text data available. I used a linear regression model as a baseline due to its simple implementation. I then implemented a random forest regression model in order to capture the non-linear relationship between the text content of the post and the number of upvotes it would receive. Finally, I used a neural network model that would further improve on the random forest regression model by tuning hyperparameters and incorporating deep learning.

\quad In the future, different models will be able to achieve lower loss in the prediction of popularity. Recurrent Neural Networks(RNN) and BERT by Google are two models that can be implemented with this task due to their applications in NLP fields. Additionally, using a larger quantity of data or more recent data would yield more accurate results, which can be achieved by scraping public data. The current neural network model can also be improved by implementing a different architecture, with the inclusion of Long-Short Term Memory(LSTM) layers or Gated Recurrent Units(GRU). 

\section{Acknowledgements}
\begin{center}
    Special thanks to mentor Jonathan Mak for giving invaluable insight into neural network techniques and constructive paper revision. Additional thanks to mentor Mayuka Sarukkai for helping during the revision process.
\end{center}

\bibliographystyle{unsrt}  


\begin{thebibliography}{1}

\bibitem{wolfsfeld 2013}
E. S. Gadi Wolfsfeld, “Social Media and the Arab Spring: Politics Comes First - Gadi Wolfsfeld, Elad Segev, Tamir Sheafer, 2013,” SAGE Journals, 16-Jan-2013. [Online]. Available: https://journals.sagepub.com/doi/full/10.1177/1940161212471716. [Accessed: 06-Jun-2021].

\bibitem{schoen2013}
H. Schoen, D. Gayo-Avello, P. T. Metaxas, E. Mustafaraj, M. Strohmaier, and P. Gloor, “The Power of Prediction with Social Media,” Wellesley College Digital Repository, 2013 [Online]. Available: https://repository.wellesley.edu/object/ir124. [Accessed: 06-Jun-2021].

\bibitem{segall2012}
J. Segall and A. Zamoshchin, “[PDF] PREDICTING REDDIT POST: Semantic Scholar,” [PDF] PREDICTING REDDIT POST | Semantic Scholar, 2012. [Online]. Available: https://www.semanticscholar.org/paper/PREDICTING-REDDIT-POST-Segall-Zamoshchin/e0733561a282d4f4c07f11a3269d9c35a446f937. [Accessed: 06-Jun-2021].

\bibitem{deaton2017}
S. Deaton, S. Hutchison, and S. J. Matthews, “Using Machine Learning to Predict the Popularity of Reddit Comments,” seandeaton.com, 2017. [Online]. Available: https://seandeaton.com/publications/reddit-paper.pdf. [Accessed: 06-Jun-2021].

\bibitem{yu2012}
S. Yu and S. Kak, “A Survey of Prediction Using Social Media,” arXiv.org, 07-Mar-2012. [Online]. Available: https://arxiv.org/abs/1203.1647. [Accessed: 06-Jun-2021].

\bibitem{muzio2021}
T. Di Muzio, “GameStop Capitalism. Wall Street vs. The Reddit Rally (Part I),” EconStor, 05-Feb-2021. [Online]. Available: https://www.econstor.eu/handle/10419/229951. [Accessed: 06-Jun-2021].

\bibitem{hasso2021}
T. Hasso, D. Müller, M. Pelster, and S. Warkulat, “Who Participated in the GameStop Frenzy? Evidence from Brokerage Accounts,” SSRN, 25-Feb-2021. [Online]. Available: https://papers.ssrn.com/sol3/papers.cfm?abstract\_id=3792095. [Accessed: 06-Jun-2021].

\bibitem{rohlin2016}
T. Rohlin, “Popularity Prediction of Reddit Texts,” SJSU ScholarWorks, 2016. [Online]. Available: https://scholarworks.sjsu.edu/etd\_theses/4704/. [Accessed: 06-Jun-2021].

\bibitem{lyocsa2021}
Š. Lyócsa, E. Baumöhl, and T. Vŷrost, “YOLO trading: Riding with the herd during the GameStop episode,” EconStor, 2021. [Online]. Available: https://www.econstor.eu/handle/10419/230679. [Accessed: 06-Jun-2021]. 

\end{thebibliography}

\end{document}